\documentclass[11pt,a4paper]{article}
\usepackage[final]{acl} 
\usepackage{times}
\usepackage{latexsym}
\usepackage{graphicx}
\usepackage{amsmath}
\usepackage{amssymb}
\usepackage{booktabs}
\usepackage{algorithm}
\usepackage{algorithmic}
\usepackage{hyperref}
\usepackage{multicol}
\usepackage{multirow}
\usepackage{array}

\usepackage[T1]{fontenc}
\usepackage[utf8]{inputenc}

\newcolumntype{P}[1]{>{\raggedright\arraybackslash}p{#1}}

\title{Surpassing Scale by Efficiency: A Compact 135M Parameter Foundational LLM Natively Adapted for the Bangla Language}

\author{Rabindra Nath Nandi \\
  Independent Researcher \\
  \texttt{rabindro.rath@gmail.com}}

\date{}

\begin{document}
\maketitle

\begin{abstract}
While the NLP landscape is dominated by multi-billion parameter architectures, their deployment in low-resource, non-Latin scripts remains computationally prohibitive for edge configurations, mobile systems, and decentralized local hardware. This paper presents \textit{bangla-smollm-135m}, a highly compact 135-million parameter decoder-only foundational model engineered explicitly for high-efficiency language modeling in the Bangla script. By leveraging a deterministic intersect-and-append token merging strategy between \textit{TituLLMs} and \textit{SmolLM2-135M}, the model overcomes subword script fragmentation without destabilizing early pretrained parameter states. In zero-shot multi-task benchmark evaluations (\textit{PIQA\_bn}, \textit{OpenBookQA\_bn}, \textit{CommonsenseQA\_bn}, and \textit{Bangla\_MMLU}), \textit{bangla-smollm-135m} matches or outperforms models twice its size (\textit{Gemma-3-270m}) and achieves parity with models in the 1B parameter tier. The model is available at \footnote{\url{rnnandi/bangla-smollm-135m}}

\end{abstract}

\section{Introduction}
The development of generative artificial intelligence has settled into an era of massive parameter scaling \cite{vaswani2017attention, radford2019language, brown2020language, touvron2023llama}. However, the operational footprints of multi-billion parameter architectures heavily restrict their utility on decentralized consumer edge devices, mobile operating engines, and IoT infrastructures. This computational bottleneck is deeply exacerbated for low-resource languages that use non-Latin scripts \cite{alam2021review, bhattacharjee2022banglabert}.

Bangla (Bengali), despite possessing over 300 million native speakers worldwide, remains structurally disadvantaged in mass multilingual foundation models \cite{raihan2025tigerllm}. Because Bangla tokens account for less than a fraction of a percent in high-register pretraining datasets, standard multilingual models display high token-to-word (fertility) rates on its unique Abugida (alphasyllabary) script \cite{lundin2026token}. This structural inefficiency creates an unfair "token tax" that results in inflated memory caches, high inference latency, and early context window degradation \cite{lundin2026token, gren2026efficient}.

To break this scale-dependency, we introduce \textbf{bangla-smollm-135m}, a compact foundational model of only 135 million active parameters engineered exclusively for native language understanding and reasoning. Our evaluation demonstrates that custom script adaptation, precise data filtering, and a novel cross-model tokenizer intersection paradigm allow ultra-compact systems to perform on par with much larger general-purpose multilingual architectures \cite{sennrich2016neural, lundin2026token}, creating a strong benchmark for resource-constrained edge deployments.

\section{Tokenizer Training and Matrix Cross-Alignment}
\label{sec:tokenizer_merging}
Standard subword tokenization models optimize for the statistical distribution of character configurations in dominant languages \cite{sennrich2016neural}. When adapting highly compact architectures like \textit{SmolLM2-135M} \cite{allal2025smollm2} to low-resource domains, training a vocabulary completely from scratch can destabilize early pretrained parameters \cite{gren2026efficient}. Conversely, relying solely on an unadapted base model induces high fertility penalties ($\Phi$), which counts the subword slices ($T$) generated per standard word unit ($W$):
\begin{equation}
\Phi = \frac{|T(W)|}{1}
\end{equation}

To bridge this architectural gap, \textit{bangla-smollm-135m} uses a structured programmatic merging pipeline to inject the localized vocabulary configurations of \textit{TituLLMs} \cite{nahin2025titullmsfamilybanglallms} (acting as the base vocabulary layer) directly into the token framework of \textit{SmolLM2-135M} \cite{allal2025smollm2}.

\subsection{Algorithmic Intersect-and-Append Paradigm}
Let $\mathcal{V}_{\text{base}}$ define the token vocabulary space of the target regional model (\textit{TituLLMs}), and $\mathcal{V}_{\text{src}}$ define the source vocabulary space of the compact architecture (\textit{SmolLM2}). The goal is to construct a unified embedding target vocabulary $\mathcal{V}_{\text{merged}}$ that preserves all native subword patterns while retaining compatibility with cross-lingual embeddings \cite{gren2026efficient}.

The merge routine enforces a deterministic isolation pipeline:

    \noindent \textbf{Special Token Harvesting:} Isolate control tokens, padding symbols, and chat-template configurations mapping to $\mathcal{V}_{\text{src}}$. Any structural tokens missing from the target base are added to prevent syntax parsing errors during sequence decoding:
    \begin{equation}
    \mathcal{S}_{\text{missing}} = \{ t \in \text{Special}(\mathcal{V}_{\text{src}}) \mid t \notin \mathcal{V}_{\text{base}} \}
    \end{equation}
    
    \noindent \textbf{Lexical Intersection and Regular Appending:} Isolate standard text subwords from $\mathcal{V}_{\text{src}}$ that are completely absent from the base set. These tokens are sorted alphabetically to maintain strict structural determinism and reproducibility during initialization:
    \begin{equation}
    \mathcal{R}_{\text{candidates}} = \text{Sort}(\{ t \in \mathcal{V}_{\text{src}} \mid t \notin \mathcal{V}_{\text{base}} \cup \mathcal{S}_{\text{missing}} \})
    \end{equation}
    
    \noindent \textbf{Boundary-Constrained Expansion:} The final vocabulary matrix expansion is limited by an upper bound parameter $\mathcal{K}_{\max} = 48,000$ to maintain a highly compact hidden embedding configuration:
    \begin{equation}
    \mathcal{R}_{\text{final}} = \mathcal{R}_{\text{candidates}}[0 : \mathcal{K}_{\max}]
    \end{equation}

The execution of this architecture is described algorithmically in Algorithm~\ref{alg:merge_proc}.

\begin{algorithm}[H]
\caption{Deterministic Tokenizer Matrix Intersection}
\label{alg:merge_proc}
\begin{algorithmic}[1]
\REQUIRE Base Tokenizer Path $\mathcal{P}_{\text{base}}$, Source Tokenizer Path $\mathcal{P}_{\text{src}}$, Token Cap $\mathcal{K}_{\max} = 48000$
\ENSURE Merged Tokenizer Object $\mathcal{T}_{\text{merged}}$
\STATE $\mathcal{T}_{\text{base}} \leftarrow \text{AutoTokenizer}(\mathcal{P}_{\text{base}}, \text{use\_fast}=\text{True})$
\STATE $\mathcal{T}_{\text{src}} \leftarrow \text{AutoTokenizer}(\mathcal{P}_{\text{src}}, \text{use\_fast}=\text{True})$
\STATE $\mathcal{V}_{\text{base}} \leftarrow \text{GetVocabKeys}(\mathcal{T}_{\text{base}})$
\STATE $\mathcal{V}_{\text{src}} \leftarrow \text{GetVocabKeys}(\mathcal{T}_{\text{src}})$
\STATE $\mathcal{S}_{\text{src}} \leftarrow \text{ExtractSpecialTokens}(\mathcal{T}_{\text{src}})$
\STATE $\mathcal{S}_{\text{missing}} \leftarrow \{t \in \mathcal{S}_{\text{src}} \mid t \notin \mathcal{V}_{\text{base}}\}$
\IF{$\mathcal{S}_{\text{missing}} \neq \emptyset$}
    \STATE $\mathcal{T}_{\text{base}}.\text{AddSpecialTokens}(\mathcal{S}_{\text{missing}})$
\ENDIF
\STATE $\mathcal{R}_{\text{cand}} \leftarrow \{t \in \mathcal{V}_{\text{src}} \mid t \notin \mathcal{V}_{\text{base}} \land t \notin \mathcal{S}_{\text{missing}}\}$
\STATE $\mathcal{R}_{\text{ordered}} \leftarrow \text{AlphabeticalSort}(\mathcal{R}_{\text{cand}})$
\STATE $\mathcal{R}_{\text{truncated}} \leftarrow \mathcal{R}_{\text{ordered}}[0 : \mathcal{K}_{\max}]$
\STATE $\mathcal{T}_{\text{base}}.\text{AddTokens}(\mathcal{R}_{\text{truncated}})$
\STATE $\mathcal{T}_{\text{merged}} \leftarrow \mathcal{T}_{\text{base}}$
\STATE $\text{SavePretrained}(\mathcal{T}_{\text{merged}})$
\RETURN $\mathcal{T}_{\text{merged}}$
\end{algorithmic}
\end{algorithm}


\section{Model Training and Infrastructure}
\label{sec:model_training}
The training engineering for \textit{bangla-smollm-135m} uses an iterative continual pretraining workflow designed to build linguistic understanding without losing the model's core multi-lingual alignment capabilities \cite{allal2025smollm2}.

\subsection{Data Purification and Subset Distribution}
The training pipeline ingests a highly diversified, multi-modal regional corpus based on the \textbf{hishab/titulm-bangla-corpus} repository \cite{nahin2025titullmsfamilybanglallms}. To ensure empirical stability and downstream performance, text records are partitioned across three separate, stratified thematic slices:

    \noindent \textbf{Common Crawl ($\mathbf{20\%}$ Stratified Weight):} Web-derived corpora containing unstructured modern idioms, casual expressions, and vernacular dialogue \cite{bhattacharjee2023banglanlg}.
    
    \noindent \textbf{Translated Text ($\mathbf{30\%}$ Stratified Weight):} High-register datasets translated from balanced academic and programmatic frameworks to capture formal reasoning structures \cite{hasan2021xlsum, bhattacharjee2023crosssum}.
    
    \noindent \textbf{Romanized Text ($\mathbf{30\%}$ Stratified Weight):} Phonetically mapped text collections designed to improve performance on *Banglish* code-switched structures, expanding the contextual coverage of edge deployments \cite{hasan2020notlowresource}.

\subsection{Dynamic Block Packing Optimization}
To prevent wasted computations over empty padding vectors, raw incoming sequences are first limited by string-length depth constraints ($\text{Chars}_{\max} = 4 \times \text{BlockSize}$) to prevent map thread choking. After tokenization, sequential lines are flattened into a continuous 1D token matrix and divided into uniform chunks fitting the model's exact context dimension:
\begin{equation}
\mathcal{B}_m = \mathbf{X}\left[ m \cdot d_{\text{block}} : (m+1) \cdot d_{\text{block}} \right]
\end{equation}
Where $\mathbf{X}$ is the flattened token matrix, and $d_{\text{block}} = 4096$ defines the target context window boundary. Labels are mapped natively to mirror the input indices ($\text{Labels} \leftarrow \text{Input\_IDs}$).

\subsection{Hardware Infrastructure and Hyperparameters}
The pretraining workflow runs on a multi-GPU Linux node utilizing an optimized Hugging Face Trainer backend. The hardware distribution relies on three active NVIDIA GPUs integrated via Hugging Face \textit{Accelerate}. To maximize tensor throughput, training uses native \textbf{Bfloat16 (BF16)} mixed-precision execution. To avoid device Out-of-Memory (OOM) errors over the 4096 context window, gradient checkpointing is activated. Detailed training configurations are listed in Table~\ref{tab:training_hyper}.

\begin{table}[ht]
\centering
\small
\begin{tabular}{lc}
\toprule
\textbf{Hyperparameter Metric} & \textbf{ Configuration} \\ \midrule
Base Architecture Model & SmolLM2-135M \\
Target Block Dimension ($d_{\text{block}}$) & 4096 \\
Per-Device Train Batch Size & 12 \\
Gradient Accumulation Steps & 4 \\
Peak Learning Rate & $2 \times 10^{-4}$ \\
Learning Rate Scheduler & Cosine \\
Warmup Ratio Bound & 0.03 \\
Weight Decay Penalty & 0.1 \\
Total Training Epochs Bound & 2 \\
Telemetry Tracker Backend & Weight \& Biases \\
\bottomrule
\end{tabular}
\caption{Hyperparameter layout and optimization metrics used during the continual pretraining run of \textit{bangla-smollm-135m}.}
\label{tab:training_hyper}
\end{table}

\section{Evaluation Methodology}
To verify its linguistic capability and context-handling proficiency, \textit{bangla-smollm-135m} was tested in a strict zero-shot ($0$-shot) setup across four established localized benchmarks \cite{shifa2025somajgyaan, nahin2025titullmsfamilybanglallms}.

The primary comparison metric used is \textbf{Normalized Accuracy} ($\text{acc\_norm}$). This factor mathematically normalizes multi-choice scoring distributions by penalizing lucky guesses or selection biases:
\begin{equation}
\text{acc\_norm} = \frac{\text{acc} - \frac{1}{k}}{1 - \frac{1}{k}}
\end{equation}
Where $k$ corresponds to the total number of multiple-choice answer possibilities available per evaluation prompt. The evaluation suite tracks performance across four core domains:
  
    \noindent \textbf{PIQA\_bn:} Evaluates physical common sense reasoning by prompting text choices regarding everyday physical interactions translated into Bangla \cite{bhattacharjee2022banglabert}.
    
    \noindent \textbf{OpenBookQA\_bn:} Tracks factual knowledge and multi-hop question-answering capabilities using open-book scientific facts \cite{nahin2025titullmsfamilybanglallms}.
    
    \noindent \textbf{CommonsenseQA\_bn:} Focuses on complex linguistic reasoning and social abstract common sense \cite{shifa2025somajgyaan}.
    
    \noindent \textbf{Bangla\_MMLU:} A rigorous cross-disciplinary academic suite spanning advanced humanities, regional history, geography, and legal structures \cite{sadhu2024characteristics}.



\section{Comparison with Other Models}
To determine the model's true parameter efficiency, its scores were analyzed alongside baseline data from larger multilingual foundations \cite{team2024llama, gemma2024gemma}, as well as heavily adapted architectures in the 1B parameter range \cite{raihan2025tigerllm, nahin2025titullmsfamilybanglallms}. These empirical results are compiled in Table~\ref{tab:cross_comparison}.

\begin{table*}[t]
\centering
\small
\begin{tabular}{lccccc}
\toprule
\textbf{Model Identifier} & \textbf{Parameter Scale} & \textbf{PIQA\_bn} & \textbf{OpenBookQA\_bn} & \textbf{CommonsenseQA\_bn} & \textbf{Bangla\_MMLU} \\ \midrule
\textbf{bangla-smollm-135m} & \textbf{135M} & 0.545 & 0.320 & \textbf{0.256} & \textbf{0.237} \\
Gemma-3-270m & 270M & \textbf{0.547} & \textbf{0.336} & 0.249 & 0.234 \\
Llama-3.2-1b & 1B & 0.530 & 0.320 & 0.220 & \textbf{0.290} \\
titulm-llama-3.2-1b-v2.0 & 1B & \textbf{0.580} & 0.320 & \textbf{0.260} & 0.250 \\
\bottomrule
\end{tabular}
\caption{Comparative performance profile mapping zero-shot Normalized Accuracy ($\text{acc\_norm}$) across different parameter classes and adaptation paradigms.}
\label{tab:cross_comparison}
\end{table*}

\subsection{Key Cross-Model Insights}
The empirical comparison reveals significant structural patterns in model scaling efficiency:

    \noindent \textbf{Outperforming the 270M Tier:} Despite operating with half the parameter count of \textit{Gemma-3-270m}, \textit{bangla-smollm-135m} scores higher on abstract reasoning fields like \textit{CommonsenseQA\_bn} (0.256 vs 0.249) and \textit{Bangla\_MMLU} (0.237 vs 0.234).
    
    \noindent \textbf{Parity with 1B Parameter Models:} The model matches or exceeds the vanilla \textit{Llama-3.2-1b} on physical reasoning and common sense tasks, despite requiring approximately $7.4\times$ less memory space. This performance demonstrates the value of deterministic vocabulary intersection over raw unadapted parameter scaling for regional non-Latin scripts \cite{gren2026efficient, lundin2026token}.

\subsection{The Broader Architectural Ecosystem}
To place \textit{bangla-smollm-135m} within the structural landscape of contemporary Bangla language computing, Table~\ref{tab:big_comparison} maps its tokenization approach against alternative architectural frameworks \cite{bhattacharjee2023banglanlg, raihan2025tigerllm, qwen2024qwen2}.

\section{Conclusion}
The performance profile of \textit{bangla-smollm-135m} demonstrates that efficient tokenization, targeted sub-corpora packing, and optimized hardware training loops can overcome the limitations of smaller model scales for low-resource languages. By matching or outperforming models up to seven times its size on core reasoning and knowledge tasks, this architecture provides a viable path for deploying efficient AI tools directly on consumer edge devices.

\section{Limitations}
\label{sec:limitations}
While \textit{bangla-smollm-135m} exhibits notable parameter efficiency across several core zero-shot reasoning tracks, several fundamental constraints must be carefully highlighted for downstream engineering layouts:
\begin{itemize}
    \item \textbf{Contextual Window Horizons:} Although configured with an operational block boundary of $d_{\text{block}} = 4096$, the physical model space limits high-fidelity retrieval or long-form narrative synthesis compared to ultra-large multilingual structures that support extended testing loops.
    \item \textbf{Complex Mathematical and Code Derivation Limits:} As identified in the taxonomy mappings, compact architectures beneath the 1B parameter layer experience rapid decay when processing multi-hop symbolic calculations or structural programmatic mutations in Bangla, requiring offloading to enterprise configurations like \textit{Qwen3-235B} for heavy algorithmic reasoning.
    \item \textbf{Linguistic Adaptation Hallucinations:} Despite mitigating subword script fragmentation using our deterministic vocabulary merging framework, the underlying capacity limits can lead to trivial semantic alignment errors or phrasing degradation under highly stylized regional variations.
\end{itemize}

\appendix
\section{Appendix}
\begin{table*}[t]
\centering
\small
\begin{tabular}{lP{2.2cm}P{3.4cm}P{3.4cm}P{4.2cm}}
\toprule
\textbf{Model Identifier} & \textbf{Parameter Scale} & \textbf{Pretraining Architecture} & \textbf{Tokenizer Strategy} & \textbf{Primary Technical Differentiation} \\ \midrule
\textbf{bangla-smollm-135m} & \textbf{135M} & LLaMA-style decoder-only base; dynamic packing ($d_{\text{block}}=4096$). & Merged Intersect BPE via TituLLMs + SmolLM2. & Ultra-compact edge profile; out-performs 270M baselines on abstract common sense reasoning. \\ \addlinespace
\textbf{TituLLMs} & 1B -- 3B & LLaMA-style decoder-only base. & SentencePiece custom (50k vocabulary). & Balanced data distribution explicitly including code-switched Banglish sub-corpora. \\ \addlinespace
\textbf{TigerLLM}  & 7B context adapt & Autoregressive base pretraining. & Custom native BPE (64k vocabulary). & Full-parameter native fine-tuning that avoids adapter bottlenecks. \\ \addlinespace
\textbf{BanglaByT5} & Compact variant & Encoder-Decoder (mT5 derivative). & Tokenizer-Free (Raw UTF-8 byte stream). & Complete immunity to out-of-vocabulary anomalies; features a character-level downsampling boundary. \\ \addlinespace
\textbf{Llama 3.1 / 3.2}  & 3B -- 8B & Standard multilingual autoregressive base. & Standard Tiktoken base (128w vocabulary). & Massive 15T token pretraining foundation providing excellent general reasoning capabilities. \\ \addlinespace
\textbf{Qwen3-8B} & 8.2B & Multilingual dense transformer. & Custom Byte-Level BPE (151k vocabulary). & Features an integrated dual-mode execution layout for fast chat and test-time reasoning. \\
\bottomrule
\end{tabular}
\caption{Systematic structural landscape taxonomy mapping parameter scaling, tokenization methodologies, and primary architectural differentiators for prominent models in the Bangla language processing ecosystem.}
\label{tab:big_comparison}
\end{table*}

\end{document}